\documentclass[11pt]{article}

\usepackage[final]{acl}

\usepackage{times}
\usepackage{latexsym}
\usepackage{amsfonts} 
\usepackage{amsmath} 
\usepackage[T1]{fontenc}

\usepackage[utf8]{inputenc}

\usepackage{microtype}

\usepackage{inconsolata}

\usepackage{graphicx}
\usepackage{multirow}
\usepackage{booktabs} 
\usepackage{svg}
\usepackage{tcolorbox}
\usepackage{multicol}
\usepackage{xcolor}
\usepackage{makecell}
\usepackage{hyperref}
\usepackage{pifont}
\usepackage{algorithm}
\usepackage{algpseudocode}
\title{Key Path Identification for Resolving Knowledge Conflicts \\via SAE-based Steering}
\author{
    Wenbo Zhang\textsuperscript{1},
    Zhongxiang Sun\textsuperscript{1},
    Zhiguang Han\textsuperscript{2},
    Jun Xu\textsuperscript{1}\thanks{Corresponding author}\\
    \textsuperscript{1}Gaoling School of Artificial Intelligence, Renmin University of China
    \\\textsuperscript{2}Nanyang Technological University \\
    \texttt{\{zhangwenbo, sunzhongxiang, junxu\}@ruc.edu.cn}, \texttt{zhan010@e.ntu.edu.sg}
}

\begin{document}
\maketitle
\begin{abstract}

Sparse autoencoder (SAE)-based steering has been widely used to address knowledge conflicts by guiding LLMs to be more faithful to the contextual knowledge. Existing methods usually perform mass steering, which modifies a large batch of SAE features identified via correlation-based methods. However, due to the inaccurate correlation and the neglected feature interactions, mass steering methods fail to precisely identify the features that play the key roles in steering and introduce a large number of redundant ones, which add noise and weaken the steering effects. Our empirical studies reveal that steering only a small subset of the identified features can achieve comparable or even better performance. Motivated by this finding, 
we propose Key Path Identification (KPI), a novel method that identifies key steering features characterized by strong causal dependencies with both upstream and downstream features. From these features, KPI constructs key paths and steers through less feature modifications. 
In this way, KPI advances SAE-based steering from quantity-driven to quality-focused, offering a perspective for more precise and interpretable model editing. Experiments in RAG tasks with knowledge conflicts show that our method improves the accuracy by 18\% on average compared to the best baseline of mass steering, effectively filtering redundant features, alleviating side effects and demonstrating the core role of key paths in steering. The code is available at \url{https://github.com/Ihildu-Baggins/Key-Path-Identification}.

\end{abstract}

\section{Introduction}

Large Language Models (LLMs) acquire parametric knowledge from pre-training~\cite{touvron2023llama2openfoundation,openai2024gpt4technicalreport,deepseekai2025deepseekv3technicalreport}, which could be outdated or inherently incorrect~\cite{de2021editing,xu2024knowledge,mitchell2022memory}. Retrieval-Augmented Generation (RAG) has been widely used to address these limitations by incorporating external contextual knowledge~\cite{lewis2020retrieval}. However, when conflicts arise between parametric and contextual knowledge, referred to as knowledge conflicts~\cite{xie2023adaptive,chen2022rich,longpre-etal-2021-entity}, the model sometimes shows a preference for its inherent parametric knowledge~\cite{su2024conflictbankbenchmarkevaluatinginfluence,zhao2025analysingresidualstreamlanguage}. This tendency can result in an incorrect output that is unfaithful to the context, impairing the model's performance in RAG and other contextual understanding tasks.

Studies have been conducted to address the unfaithful outputs caused by knowledge conflicts, including instruction-based methods~\cite{wang2024resolvingknowledgeconflictslarge} and model editing methods~\cite{sun2025redeep,ferrando2025iknowentityknowledge,fayyaz2025steeringmoellmsexpert}.
Due to their lower cost, model editing methods have been popular. For example, SAEs~\cite{bricken2023monosemanticity,cunningham2023sparseautoencodershighlyinterpretable} and SAE-based steering~\cite{zhao2025steering} improve the precision and control ability of editing through monosemanticity. With a sparsity penalty applied when encoding and reconstructing the neuron representations, more monosemantic SAE features are constructed to edit and gain better steering effects.


\begin{figure}[t]
\centering
\includegraphics{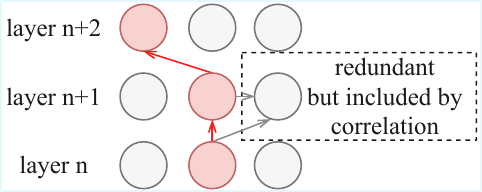}
  \caption{Strong features (red circles) may also activate (arrows) redundant weak features (gray circles), leading to the inaccurate correlation with model behaviors.}
  \label{fig:interaction}
\end{figure}

However, for the purpose of 
monosemanticity, the number of steering features in SAE-based methods could be large. For example, a SAE trained on the residual stream of a certain layer in the Gemma-2-9B model~\cite{lieberum2024gemma} has a feature dimension that is 32 times larger than the neuron dimension, reaching 131 K. 
The correlation-based identification methods (e.g., mutual information) treat features as independent units, ignoring their interactions. Consequently, a large batch of features is selected to steer, including a lot of redundant features that dilute the impact of critical features (Figure~\ref{fig:Single and Mass}).
Specifically, a feature that truly has a strong steering effect requires some features from previous layers to activate it, and they may also activate other features in the current or subsequent layers with minor steering effects themselves (Figure~\ref{fig:interaction}). 
Current SAE-based steering methods (e.g., STA~\cite{wang2025beyond}, SPARE~\cite{zhao2025steering}) apply simple and rough pruning, but still retain a large number of redundant features. 

In this work, 
we analyze the work mechanism of features, including the causal interactions among features, the enhancement of the attention scores on golden answer tokens within the context, and the gradual formation of knowledge selection behavior. Based on these analyses, we propose Key Path Identification (KPI), 
which further identifies key features and key paths that are causally critical for steering, thereby reducing noise from redundant features. 
Specifically, KPI uses a small development dataset to construct the interactions among positive features and form an interaction graph. Key features are 
characterized 
by the high indegrees, and key paths are built by the key features in the key layer with the strong steering effect and its preceding layers.


The contributions of this paper are as follows:
\begin{itemize} 
\item Our analysis reveals that SAE-based mass steering suffers from feature redundancy, impairing the steering performance.
\item 
We propose a method called Key Path Identification (KPI) which locates the key feature paths that play the core steering role, and enhances the precision and effectiveness of steering. 

\item Experiments in RAG tasks with knowledge conflicts show that KPI improves the accuracy by 18\% on average compared to the best baseline of mass steering. Empirical analysis provides mechanistic interpretations and shows that the improvements are achieved via the precise location of SAE features and the alleviation of side effects.
\end{itemize}

\begin{figure*}[!t]
\centering
\includegraphics{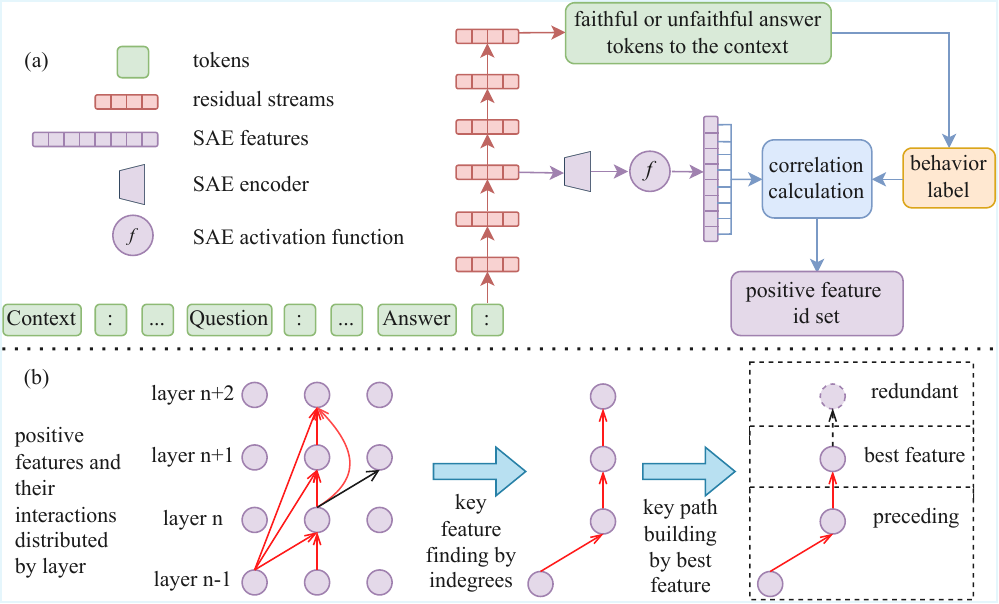}
  \caption{\textbf{The work flow of SAE-based feature selection}. (a)   shows the basic process of correlation-based identification methods of positive features. 
   (b) shows the process of our KPI method that further applies key feature finding and key path building.}
  \label{fig:structure}
\end{figure*}

\section{Preliminary}
\subsection{SAEs and SAE-based Steering}
Previous studies have shown that neurons in LLMs are multifunctional~\cite{bills2023language}, where multiple knowledge or concepts are entangled together. To get more monosemantic representations, sparse autoencoders (SAEs) are introduced~\cite{shu2025survey}. Specifically, SAEs first encode the model representation (e.g., residual streams) $\mathbf{h} \in \mathbb{R}^{d}$ into a sparser SAE representation $\mathbf{z} \in \mathbb{R}^{d_{\text{sae}}}$:
\begin{equation*}
    \mathbf{z} = \operatorname{activation\_function}\left( \mathbf{hW}_{\mathrm{enc}} + \mathbf{b}_{\mathrm{enc}} \right),
\end{equation*}
where $d_{\text{sae}} \gg d$, $\mathbf{W}_{\mathrm{enc}}$ is the encoder matrix and $\mathbf{b}_{\mathrm{enc}}$ is the bias term. Every dimension of $\mathbf{z}$ is called a SAE feature, and the
$\operatorname{activation\_function}$ can be JumpReLU~\cite{rajamanoharan2024jumpingaheadimprovingreconstruction} or TopK~\cite{bussmann2024batchtopksparseautoencoders}.  
$\mathbf{z}$ is then used to reconstruct $\mathbf{h}$:
\begin{equation*}
    \mathbf{h}_{\mathrm{sae}} = \mathbf{zW}_{\mathrm{dec}} + \mathbf{b}_{\mathrm{dec}},
\end{equation*}
where $\mathbf{h}_{\mathrm{sae}} \in \mathbb{R}^{d}$, $\mathbf{W}_{\mathrm{dec}}$ is the decoder matrix and $\mathbf{b}_{\mathrm{dec}}$ is the bias term. 
The parameters $\mathbf{W}_{\mathrm{enc}}$, $\mathbf{b}_{\mathrm{enc}}$, $\mathbf{W}_{\mathrm{dec}}$ and $\mathbf{b}_{\mathrm{dec}}$ are optimized via minimizing 
\begin{equation*}
\mathcal{L}(\mathbf{z})=\underbrace{\left\|\mathbf{h}-\mathbf{h}_{\text{ sae }}\right\|_{2}^{2}}_{\mathcal{L}_{\text{ reconstruction }}}+\underbrace{\gamma\|\mathbf{z}\|_{0}}_{\mathcal{L}_{\text{ sparsity }}},
\end{equation*}
where $\gamma \in R^+$ is a hyperparameter.

During the steering phase, given a specific SAE feature indexed by $i$, $\mathbf{W}_{\mathrm{dec}}[i,:]$ is considered as the corresponding steering vector, and the vector after steering $\mathbf{h}^{\prime}$ is calculated as
\begin{equation}
\label{steering formula}
\mathbf{h}^{\prime} = \mathbf{h} + \mathbf{\alpha W}_{\mathrm{dec}}[i,:],
\end{equation}
where $\alpha \in R^+$ is the steering strength.

\subsection{Correlation-based Identification Method of Steering Features}
The definitions in this section follow the work of SPARE~\cite{zhao2025steering}. The question answering template can be seen in Appendix~\ref{sec:Prompt Template}. We define the behavior label set $Y = \{N, T\}$, where $N$ represents the nontarget (unfaithful) behavior and $T$ represents the target (faithful) behavior. 
After sampling behavior labels of model outputs and SAE representations at the last token in the prompt, which is the closest to the generated answer, 
we choose mutual information as the correlation-based identification method of steering features. Let the random variable $Z_i$ be the activation of the specific SAE feature indexed by $i$. Mutual information between $Y$ and $Z_i$ is calculated as follows:
\begin{equation*}
I\left(Z_{i} ; Y\right)=\sum_{z_{i} \in Z_{i}} \sum_{y \in\left\{N, T\right\}} P\left(z_{i}, y\right) \log \frac{P\left(z_{i}, y\right)}{P\left(z_{i}\right) P\left(y\right)}.
\end{equation*}

By this definition, a higher mutual information suggests a stronger correlation between $Z_i$ and the behavior selection. Positive features are defined as those exhibiting higher average activations during target behavior than during nontarget behavior.

Previous method~\cite{zhao2025steering} sorts $\left\{ I\left( Z_{i} ; Y\right)\right\}_{i = 1}^{d_{sae}}$ of features in the descending order and uses $K$ as the hyperparameter to select the most correlated features:
\begin{equation}
k = \operatorname*{arg\,min}_{k} \sum_{u=1}^{k} \frac{sorted(\{I(Z_i; Y)\}_{i = 1}^{d_{sae}})_{u}}{\sum_{j=1}^{d_{sae}} I(Z_j; Y)} \geq K.
\label{eq:argmin}
\end{equation}
In this work, we work on the identified positive features. And from Equation~\ref{eq:argmin}, we can know that a simple and rough pruning is made. However, the number of selected features can still be more than 50  in a single layer to achieve ideal steering effects (Section~\ref{sec:feature-redundancy}), while steering is often performed on multiple layers. As discussed before, this pruning still includes a lot of redundant features with minor steering effects that weaken the overall performance.
More details of mutual information calculation can be seen in Appendix~\ref{sec:implementation-details}.

\section{Redundancy Comes from Inaccurate Correlation and Neglected Interactions}

\label{sec:feature-redundancy}
Since mass steering methods view the selected positive features as a whole for steering, without considering the specific work mechanism of each individual feature, this poses an obstacle to interpretability. We first try to steer each positive feature individually to observe the effect of single steering. 

\begin{figure}[t]
\includegraphics[width=\columnwidth]{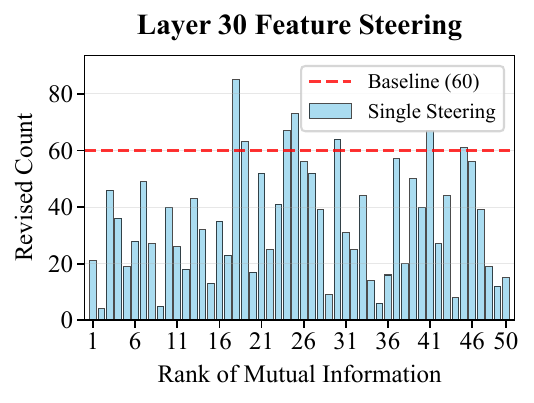}
  \includegraphics[width=\columnwidth]{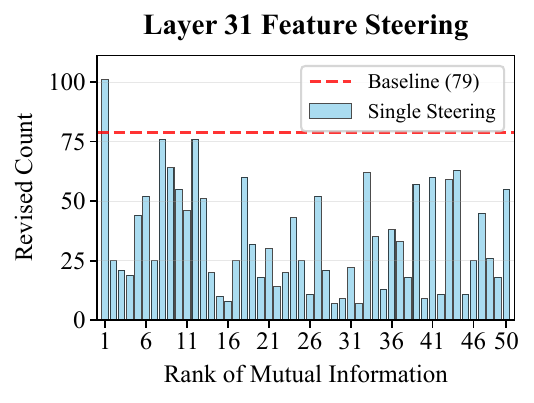}
  \caption{The performance comparison of the single feature steering and the mass steering baseline (SPARE) on layer 30 and 31 of Gemma-2-9B on a development set with the size of 200. Some single feature steerings outperform the mass steering on all the positive features.}
  \label{fig:Single and Mass}
\end{figure}

Surprisingly, we find that some features themselves have strong steering effects, which even outperform mass steering on all the positive features. As shown in Figure~\ref{fig:Single and Mass}, most features show relatively lower steering effects (the revised counts of unfaithful answers after steering), while a small number of features demonstrate significantly stronger ability. Moreover, in some layers like layer 30 of Gemma-2-9B, the weak features even rank ahead and the most useful feature has a relatively lower rank, which also shows the inaccuracy 
of the correlation. The low rank forces an increase in $K$ in Equation~\ref{eq:argmin} to include good features when using mass steering and also makes more weak features included. Due to the neglected interactions among these features, redundancy is introduced, which hinders the steering effects of the good features and thereby weakens the overall performance. This insight motivates us to not only evaluate the steering effects of individual features, but also further understand the interactions among features, how they promote or hinder each other, paving the way for obtaining a better set of steering features. 

\section{Method}
To overcome the redundancy issue in mass steering, we propose Key Path Identification (KPI), a causal dependency-driven method that identifies critical features for precise model steering. KPI operates in three stages: (1) \emph{Feature Interaction Pattern Capture}: it captures feature interaction patterns with a small development set; (2) \emph{Key Feature Finding}: it builds a feature interaction graph and identifies key features that exhibit key roles within the graph based on indegrees; (3) \emph{Key Path Building}: it builds key paths by filtering the key features following the key functional layer. This approach shifts SAE-based steering from quantity-driven to quality-focused, enhancing the steering effect. The formalization and pseudocode can be seen in Appendix~\ref{app:kpi}.

\subsection{Feature Interaction Pattern Capture}
Existing works~\cite{abnar2020quantifying,ameisen2025circuit,kamath2025tracing} construct a detailed and complex inference circuit of the model through layer-by-layer backward propagations from the output logits to observe feature interactions. However, due to the long context and the large number of features, this method also incurs high computational cost. Moreover, the constructed circuit is usually tailored for a specific input example, leading to poor generalization.

Our aim is to efficiently capture the general interaction patterns among positive features. Specifically, from each selected layer, we pick out the set of top positive features and steer each feature singly. In a single steering, for each affected layer (including the intervened feature layer and subsequent layers), we record the top 5 features with the greatest average increases in activation values. Experiments show that these interaction patterns are stable: whether on as many as 200 or as few as 10 sampled instances, the impacts of enhancing one feature on the activation increases of other features are highly consistent (see Appendix~\ref{Sensitivity}). Therefore, only a small development set is needed to efficiently and reliably capture the interaction patterns among positive features. 
The reason we do not use the increases in activation frequencies as metrics is that some features already have high baseline activation frequencies and the maximum values are 1.0, which means that the increases in activation frequencies are poor indicators of the impact degrees.

\subsection{Key Feature Finding}

The idea of using indegrees to find key features is rooted in the network bottlenecks. A high indegree shows that a node is a convergence point for multiple paths, where information flows most. We view features as nodes and their causal interactions as directed edges, which form an interaction graph of positive features.
In the graph, if a feature node has a low indegree, it may indicate that this feature is just a redundant one with a weak steering effect mistakenly identified by correlation and may hinder the performance of good features; or this feature works by activating good downstream features. If a feature node has a high indegree, it indicates that many positive features work together to promote its activation and this feature plays an important role in the mass steering effect of the positive features. Therefore, we rank the features in the descending order based on their indegrees, and the top-ranked features are considered as the key features.


\subsection{Key Path Building}
The top-ranked key features naturally form cross-layer interaction paths that play the important role in steering. However, through evaluating the single steering effect of each feature along the paths, we find that the steering effects of features have different levels.  
In some particular layers, the steering effects of the top-ranked features reach the peak—a finding that is consistent with previous studies~\cite{jin2024cutting,zhao2025steering,wang2025unveiling} that identify certain functional layers as the key to contextual understanding. 
We identify the layer containing the top-ranked feature with the best steering effect as the key layer.

As the knowledge selection behavior is performed in a range of layers, we choose to steer the identified top-ranked features in the key layer and its preceding layers, which form key paths. This is because steering only affects the current layer and subsequent layers. If we only steer the key layer, we will drop its accompanying impacts on the preceding layers, while impacts of the key layer on subsequent layers will make steering features in subsequent layers redundant. In that way, we can also deal with higher indegrees that naturally form in later layers by selecting the key layer iteratively  until no top-ranked features in preceding layers have better single steering effects.
More implementation details can be seen in Appendix~\ref{sec:implementation-details}.

\section{Experiment Setting}
\subsection{Models and SAEs}
The models and their corresponding SAEs used in our experiments are: Gemma-2-9B and SAEs from Google DeepMind~\cite{lieberum2024gemma}; Llama-3.1-8B and SAEs from Llama Scope~\cite{he2024llamascopeextractingmillions}; Llama-3-8B and SAEs from EleutherAI\footnote{https://huggingface.co/EleutherAI/sae-llama-3-8b-32x}. All the SAEs are trained on the residual streams and 131K in width. In terms of activation functions, SAEs of Gemma-2-9B and Llama-3.1-8B use JumpReLU, while SAEs of Llama-3-8B use TopK. We use Qwen3-14B~\cite{qwen3technicalreport} to judge the correctness of question answering, and the prompt template is given in Appendix~\ref{sec:Prompt Template}.

\subsection{Steering}
Since the number of our steering features is small, the steering strength needs to be large to achieve the maximal steering effect. The steering strength $\alpha$ in Formula~\ref{steering formula} is explored as a hyperparameter. We choose the largest activations of features as bases and explore their multiples to find the best steering strength more efficiently. Specifically, we use the APIs of Neuronpedia~\cite{neuronpedia} to get the largest activations of features in SAEs for Gemma-2-9B and Llama-3.1-8B, while we sample the largest activations for Llama-3-8B on a larger dataset for training. The selected features and their steering strength can be seen in Appendix~\ref{sec:implementation-details}. The steering positions are at the last token in the prompt.
\subsection{Datasets}
We choose NQ-Swap~\cite{longpre-etal-2021-entity} and Macnoise~\cite{hong-etal-2024-gullible} to evaluate the steering effect of making models more faithful. Every example in the datasets has four parts: original context, golden answer to original context, substituted context where the original golden answer is replaced, and golden answer to substituted context. Specifically, for every model, we sample the answers and identify the examples where the model still outputs the original golden answer with substituted context. A steering is considered to be successful when the model changes to output the golden answer to substituted context after steering. The feature identification methods are conducted on different training subsets of Macnoise and NQ-Swap for every model. More details can be seen in Appendix~\ref{sec:implementation-details}. Tests on another faithful dataset for generalizability can be seen in Appendix~\ref{complex-faithful}.
\subsection{Baselines}
We choose STA~\cite{bricken2023monosemanticity} and SPARE~\cite{zhao2025steering} as the SAE-based steering baselines that perform mass steering on a large number of SAE features. Also, we include ICL (in-context learning)~\cite{GPT3} as the instruction-based method, and CAD~\cite{shi-etal-2024-trusting} as a representative of the contrastive decoding methods. More implementation details and hyperparameters can be seen in Appendix~\ref{sec:implementation-details}.

Though the test datasets are selected to have unfaithful answers to the context originally, the LLM judgement has randomness during evaluation. We also report the results without steering to show the original performance of the models.
\section{Experimental Results}

\begin{table*}[t]
\centering
\small
\resizebox{\linewidth}{!}{
\begin{tabular}{lcccccc} 
\toprule
\bf Method & \multicolumn{3}{c}{\textbf{NQ-Swap}} & \multicolumn{3}{c}{\textbf{Macnoise}} \\ 
\cmidrule(lr){2-4} \cmidrule(lr){5-7}
& \bf Llama-3-8B & \bf Llama-3.1-8B & \bf Gemma-2-9B & \bf Llama-3-8B & \bf Llama-3.1-8B & \bf Gemma-2-9B \\
\midrule
Without Steering & $0.33_{\pm 0.58}$ & $0.00_{\pm 0.00}$ & $0.67_{\pm 0.29}$ & $2.33_{\pm 1.26}$ & $2.67_{\pm 1.04}$ & $2.50_{\pm 0.00}$ \\
ICL & $22.83_{\pm 1.61}$ & $16.00_{\pm 0.50}$ & $13.67_{\pm 0.58}$ & $9.83_{\pm 3.01}$ & $15.50_{\pm 1.80}$ & $7.50_{\pm 2.50}$ \\
CAD & $ 8.17_{\pm 1.04}$ & $14.67_{\pm1.89}$ & $11.50_{\pm 1.80}$ & $16.83_{\pm 4.31}$ & $14.83_{\pm 1.15}$ & $17.00_{\pm 3.04}$ \\
STA & $61.17_{\pm 2.52}$ & $24.00_{\pm 0.00}$ & $31.67_{\pm 1.26}$ & $47.17_{\pm4.25}$ & $38.33_{\pm 2.47}$ & $14.83_{\pm 1.53}$ \\
SPARE & $68.83_{\pm 2.02}$ & $48.83_{\pm 1.15}$ & $69.50_{\pm 1.50}$ &  $55.67_{\pm 2.75}$ & $54.33_{\pm 3.33}$ & $43.00_{\pm 6.54}$ \\
KPI & $\textbf{72.83}_{\pm 4.31}$ & $\textbf{50.67}_{\pm 0.76}$
 & $\textbf{79.33}_{\pm 0.76}$ & $\textbf{63.83}_{\pm 3.21}$ & $\textbf{58.50}_{\pm 5.27}$ & $\textbf{75.83}_{\pm 5.62}$  \\

\bottomrule
\end{tabular}
}
\caption{Main result table of different methods and models.}
\label{tab:main-results}
\end{table*}

\begin{figure}[t]
\includegraphics[width=\columnwidth]{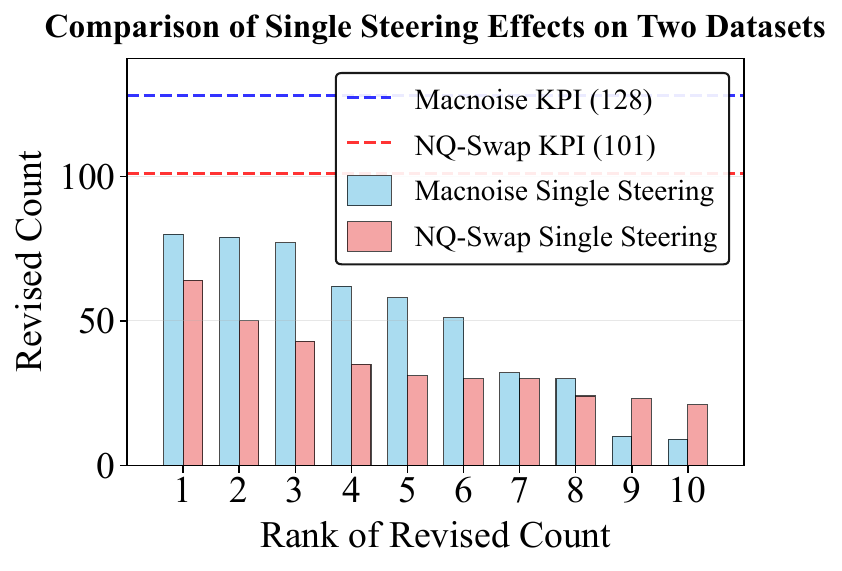}
  \caption{Comparison of single steering effects of selected key features of Llama-3.1-8B on two datasets.
  }
  \label{fig:dataset_single}
\end{figure}

\begin{figure}[t]
\includegraphics[width=\columnwidth]{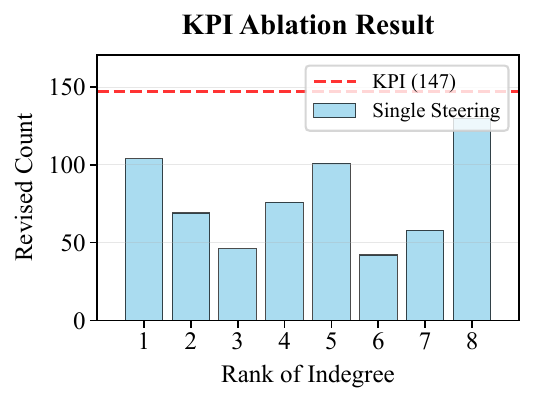}
  \caption{KPI ablation result of Gemma-2-9B. KPI surpasses every single steering.
  }
  \label{fig:Ablation}
\end{figure}

\begin{table*}[t]
\centering
\small
\resizebox{\linewidth}{!}{
\begin{tabular}{lcccccc} 
\toprule
\bf Method & \multicolumn{3}{c}{\textbf{NQ-Swap}} & \multicolumn{3}{c}{\textbf{Macnoise}} \\ 
\cmidrule(lr){2-4} \cmidrule(lr){5-7}
& \bf Llama-3-8B & \bf Llama-3.1-8B & \bf Gemma-2-9B & \bf Llama-3-8B & \bf Llama-3.1-8B & \bf Gemma-2-9B \\
\midrule
Without Steering & $99.83_{\pm 0.29}$ & $99.00_{\pm 0.50}$ & $100.00_{\pm 0.00}$ & $99.33_{\pm 0.29}$ & $100.00_{\pm 0.00}$ & $99.50_{\pm 0.50}$ \\
SPARE & $88.17_{\pm 4.01}$ & $84.50_{\pm 1.80}$ & $76.83_{\pm 4.25}$ & $82.83_{\pm 3.69}$ & $82.00_{\pm 0.87}$ & $95.33_{\pm 1.89}$ \\
KPI & $95.00_{\pm 0.50}$ & $98.00_{\pm 0.87}$ & $99.17_{\pm 0.58}$ & $95.17_{\pm 1.44}$ & $96.67_{\pm 1.44}$ & $95.50_{\pm 0.50}$ \\

\bottomrule
\end{tabular}
}
\caption{Side effect checks of our method on faithful examples. Our method achieves better or comparable performance to the baseline, showing the ability to alleviate side effects by filtering redundant features.}
\label{tab:side-results}
\end{table*}

\begin{figure*}[htbp]
   \centering
   \begin{minipage}[t]{0.48\linewidth}
       \centering
       \includegraphics[width=\linewidth]{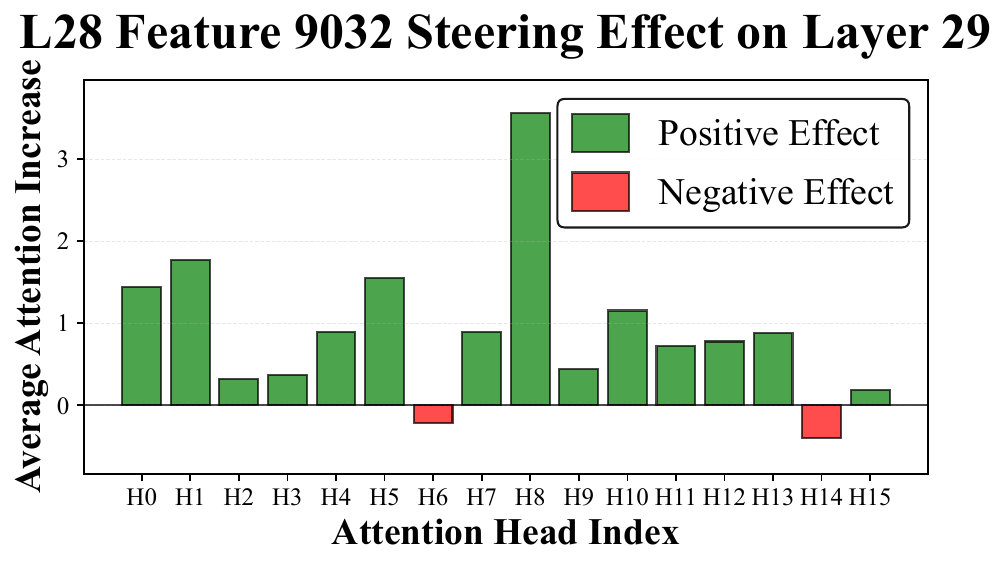}
       \caption{Impact of the key feature on the next key layer's attention score changes.}
       \label{fig:key_feature_key_layer_att}
   \end{minipage}
   \hfill
   \begin{minipage}[t]{0.48\linewidth}
       \centering
       \includegraphics[width=\linewidth]{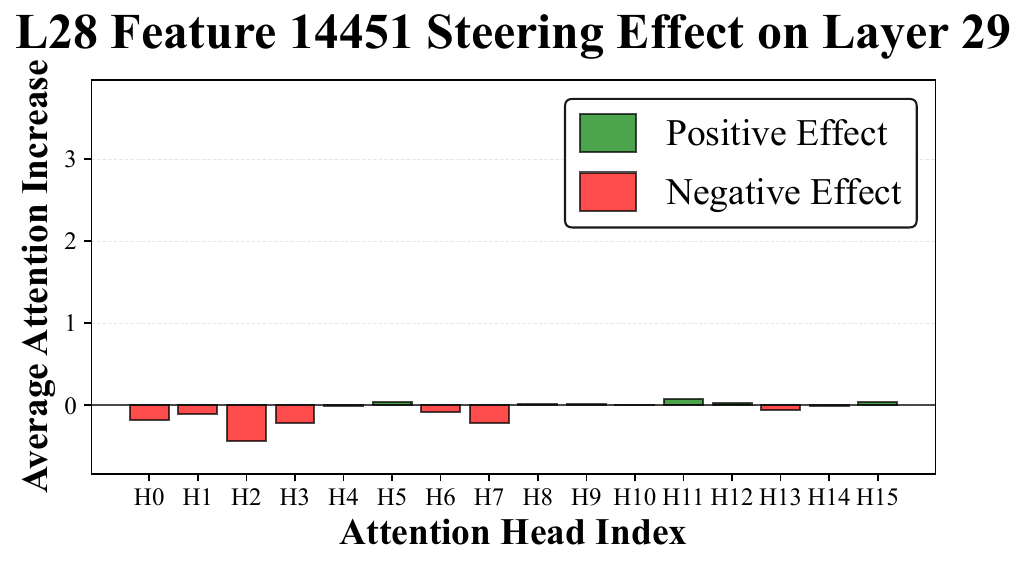}
       \caption{Impact of a feature with relatively high mutual information score but a weak steering effect on the next key layer's attention score changes.}
       \label{fig:weak_feature_key_layer_att}
   \end{minipage}
    \hfill
      \begin{minipage}[t]{0.48\linewidth}
       \centering
       \includegraphics[width=\linewidth]{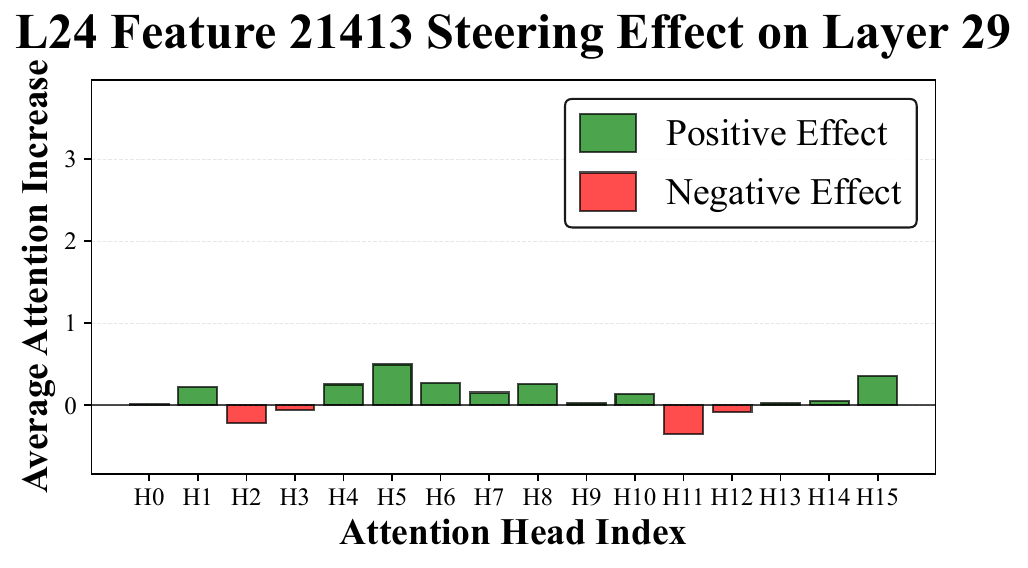}
       \caption{Impact of a feature with relatively strong steering effect on the next key layer's attention score changes.}
       \label{fig:strong_feature_key_layer_att}
   \end{minipage}
    \hfill
      \begin{minipage}[t]{0.48\linewidth}
       \centering
       \includegraphics[width=\linewidth]{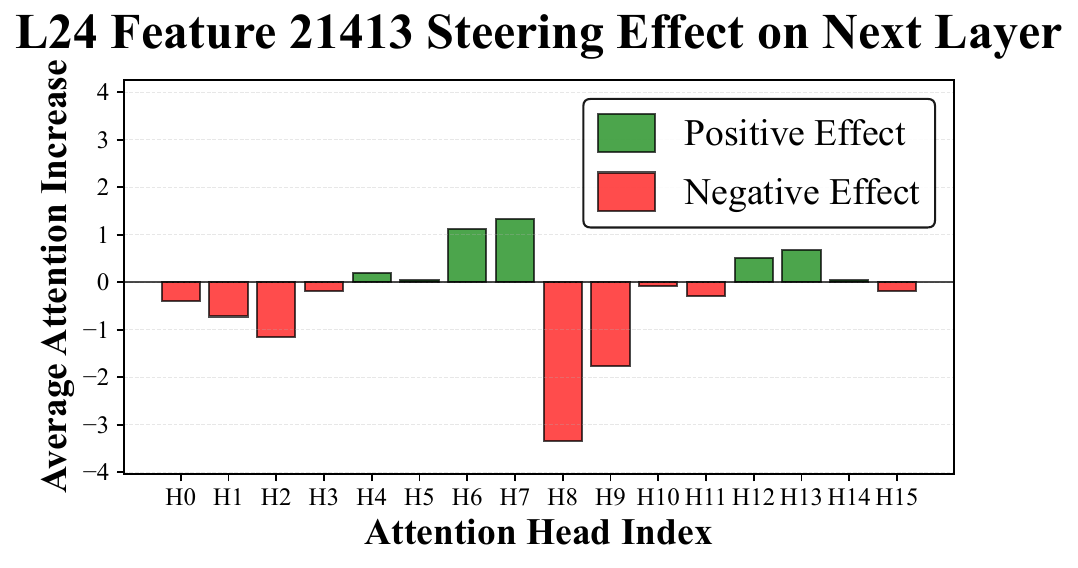}
       \caption{Impact of a feature with relatively strong steering effect on the next layer's attention score changes.}
       \label{fig:strong_feature_next_layer_att}
   \end{minipage}
\end{figure*}

\begin{figure}[t]
\includegraphics[width=\columnwidth]{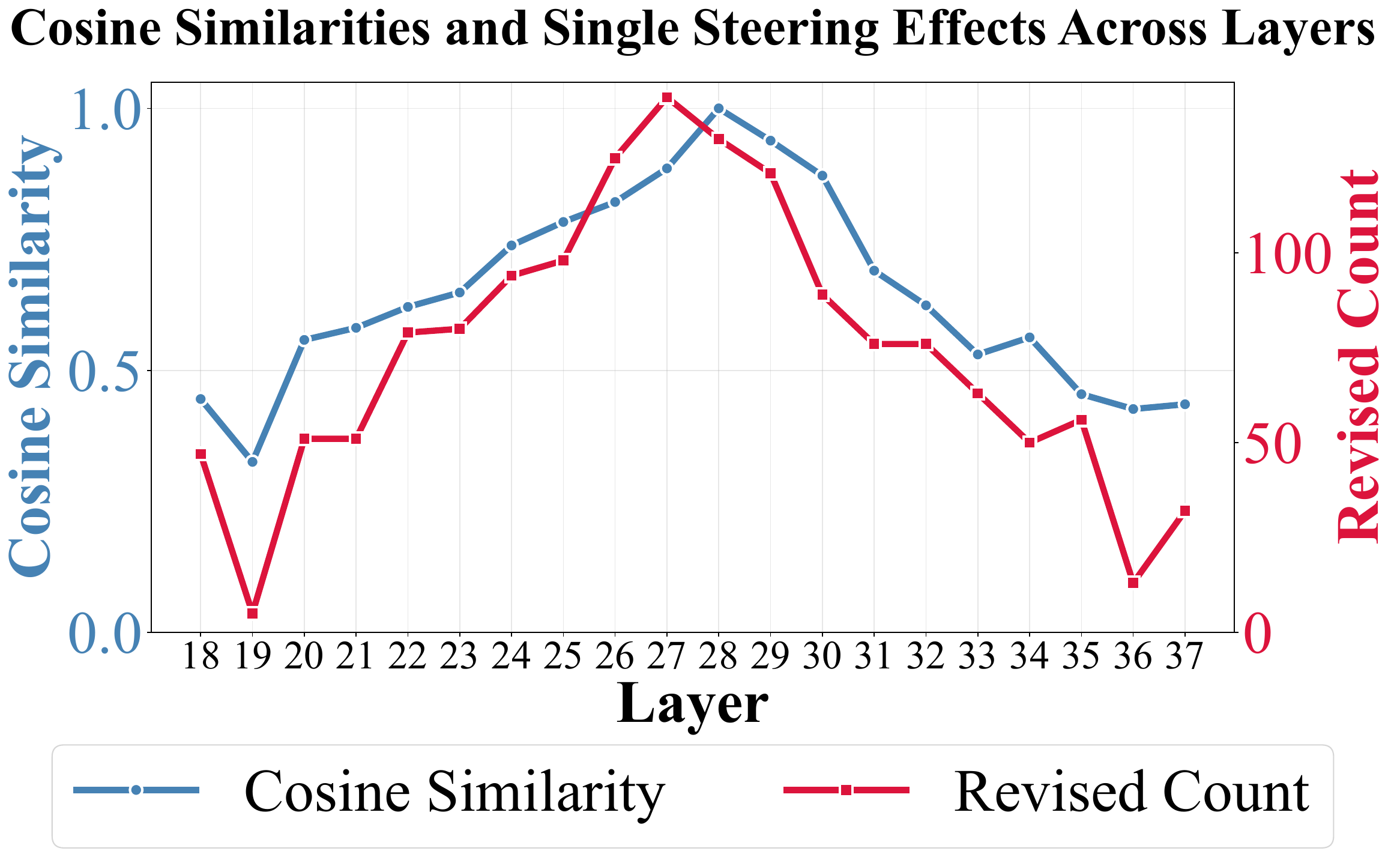}
  \caption{The cosine similarities and single steering effects of the most similar features across the layers of the key feature (located in layer 28).
  }
  \label{fig:cos}
\end{figure}

\subsection{Main Results}
We present the main results (accuracies) in Table~\ref{tab:main-results}. Our KPI method achieves the best performance on both the datasets for every model. 

For the Llama-3.1-8B model, SAE-based methods show lower performance on NQ-Swap. We check the single steering results and find that single steering shows the same decreasing tendency (Figure~\ref{fig:dataset_single}), and KPI still outperforms single steering. It indicates that the overall drop originates from the effect drops of features, showing that the SAE-based steering methods rely on the quality of SAEs. 
Another issue is that the performance of SPARE for Gemma-2-9B on Macnoise is relatively low due to inaccurate layer selection. Though SPARE and KPI both search for a wide range of layers, SPARE cannot precisely locate the important layers. Because mass steering cannot precisely reflect the effectiveness of layers as shown in our Section~\ref{sec:feature-redundancy}, which seriously hinders its effect. Our method achieves better key layer location through selecting the key layer iteratively until there are no better top-ranked features in preceding layers.

\subsection{Ablation Result}
To verify that our method filters the redundant features and finds the features that promote the overall effect rather than weaken the good features, we check the single steering effects of the features selected by our method.
As shown in Figure~\ref{fig:Ablation}, the number of selected features is only 8, which is much smaller than those of the mass steering baselines that steer more features even in one layer. Moreover, the average single steering effect is much better than that in Figure~\ref{fig:Single and Mass}, which indicates that the redundant features are mostly excluded. And the overall effect of KPI is better than every single steering, showing that the features collaborate to achieve better results. As our method includes the upstream features of the key layer and gains benefits, it shows that the model performs the knowledge selection behavior in multiple layers and steering on preceding layers is necessary.

\subsection{Side Effect Checks}
As the editing inevitably influences the usage of parametric knowledge, we conduct side effect checks of our steering. Since our test datasets of main results are unfaithful examples originally, we check the effect of our method on the faithful examples with no knowledge conflicts to see the side effects.

The results can be seen in Table~\ref{tab:side-results}. Our method achieves comparable or better performance, showing significant improvement in faithfulness and the relatively small side effects. More importantly, the comparison with the mass steering baseline shows that our method successfully filters redundant features that bring noise, which also impairs the effects of good features and models' ability. Side effect checks of other downstream model capabilities can be seen in Appendix~\ref{Other Model Capabilities}.

\subsection{Attention Score Influence Result}
To further understand how features work when steering, we sample the attention score changes on golden answer tokens within the context of single steering. As the SAEs are trained on the post residual streams, the steering influences the attention scores of subsequent layers. We perform the sampling on the layer next steering and the next key layer. Here we report the results of Gemma-2-9B from Figure~\ref{fig:key_feature_key_layer_att} to Figure~\ref{fig:strong_feature_next_layer_att}.

From the results, we find that the recognized key features (a representative located in layer 28 shown in Figure~\ref{fig:key_feature_key_layer_att}) and many strong features (a representative located in layer 24 shown in Figure~\ref{fig:strong_feature_key_layer_att}) consistently enhance the attention scores on golden answer tokens of most attention heads in the next key layer. For comparison, the result of a feature with a relatively high mutual information score but a weak steering effect is shown in Figure~\ref{fig:weak_feature_key_layer_att}, which has lower enhancements and more declines, again showing the inaccuracy of the correlation.

We observe that some features work by activating the good features and the existence of functional layers. Although many strong features have overall positive attention effects on the next key layer (shown in Figure~\ref{fig:strong_feature_key_layer_att}), they show obviously more negative attention effects on the adjacent next layer (shown in Figure~\ref{fig:strong_feature_next_layer_att}). This kind of features does not make the model focus on the golden answer tokens immediately, but activates the good features for higher activations through model inferences and finally enhances attentions on the answers in subsequent layers that are more correlated with knowledge conflicts and behavior selection. 
It also indicates that these features are located in the non-functional layers that are distant from the layers that perform the target function. This inspires us to analyze how this activating process works. 

\subsection{Cosine Similarity Result}
As the SAE-based steering uses the decoding vectors to change the residual streams, we can view the decoding vectors as the representations of the features and their activation patterns in the vector form. Based on this idea, we can calculate the cosine similarities of the decoding vectors to evaluate the way of interactions among the features.

Here we present the cosine similarities and single steering effects of the most similar features across the layers of the key feature (located in layer 28 of Gemma-2-9B) (Figure~\ref{fig:cos}). We can see that there are features with high vector cosine similarities in the adjacent layers. Overall, both the similarities and steering effects have the tendency of diminishing with distance from the key layer. Moreover, if we add the decoding vector of the key feature to the residual streams in earlier layers, it can still perform a comparable strong steering effect to that when steering in the original layer. It shows that these highly similar features work because they directly add the key feature patterns into the residual streams. 
In terms of the features identified by correlation in further layers, as the patterns are more different, they activate the key feature through more natural and complex computations. 
From the results of Figure~\ref{fig:cos}, we can estimate that the patterns represented by this key feature mainly form and work during the 26-29 layers. It shows that the patterns of behavior selection gradually form and function in a range of layers, which is also evidence of the existence of functional layers, verifying the benefit of steering on preceding layers.



\section{Conclusion}
This work presents Key Path Identification (KPI) for precise model steering, addressing knowledge conflicts in RAG tasks. The key findings include: (1) we diagnose the critical limitation of existing SAE-based mass steering methods: feature redundancy caused by the inaccurate correlation and the neglected feature interactions; (2) we introduce KPI which identifies critical features based on causal dependencies, establishing a new perspective for precise and interpretable model editing with less feature modifications; (3) we provide empirical validations that show KPI's superiority, enhancing feature collaboration to improve the overall performance, effectively filtering redundant features, and alleviating side effects. Analysis on attentions and feature similarities provides mechanistic interpretations, demonstrating that steering works by enhancing attentions to the golden answer tokens and that knowledge selection behavior gradually forms in multiple layers. By prioritizing quality over quantity, KPI advances SAE-based mass steering, offering an effective and interpretable solution for improving LLM faithfulness.

\section*{Limitations}
\label{limitation}
\paragraph{Basic Methods}
Correlation-based methods have their inherent drawbacks with their nature of acausality, and our method tries to reduce the noise by utilizing the causal interactions among key features. Meanwhile, the circuit finding methods of backward propagations are not appropriate for long context analysis, which require pruning that may also risk losing causal features. More investigations should be done to optimize the identification methods fundamentally. Some works~\cite{lu2026assistant,wang2025enhancing} try to denoise from the perspective of data filtering, which is not deeply investigated in this work.
\paragraph{Multiple Circuits}
Some studies have found that there are multiple computation circuits that have similar functions~\cite{lindsey2025biology}, which is an embodiment of the model's robustness. 
For example, we find a circuit that enables the model to solve addition problems, and then disable it by making the activations of features in the circuit to zeros, but the model can still perform additions through other circuits that can also enable the model to do it. 
Our method finds the simple circuit that is with the most correlation, but does not exclude the existence of backup circuits. 
Moreover, it's worth noting that activating multiple circuits with the same function may not bring benefits, because of the saturation of ability and mutual interference.
\paragraph{Models and SAEs}
As presented above, our method requires cross-layer analysis. So the SAEs trained on continuous layers are needed. However, we find that no satisfying SAE resources are available for larger sizes (70B+) of models. In Neuronpedia, which is the largest SAE provider, there are only SAEs in limited layer settings for the models in such sizes. And training SAEs from scratch for such large models is too costly. We still emphasize that we try our best in the model selection for generalizability. Two model families and two SAE structures are included. And in other works~\cite{lu2026assistant} that simply use the differences of residual streams between the positive and negative examples, they also show the generalizability in larger models. The good generalizability of basic steering methods shows the tendency that our denoising methods can work as well, because they have the same basic mechanism.
\paragraph{RAG Setting}
In realistic RAG applications, many additional conditions may arise. In our experiments, we assume that the retrieved evidence should be followed, and we focus on improving faithfulness to context rather than broader reasoning ability.

\section*{Acknowledgements}
This work was funded by the National Natural Science Foundation of China (62472426). Work partially done at Beijing Key Laboratory of Research on Large Models and Intelligent Governance, and Engineering Research Center of Next-Generation Intelligent
Search and Recommendation, Ministry of Education.

\bibliography{custom}

\appendix

\section{Related Work}
\subsection{Mechanistic Interpretability}
Previous studies on interpretability of LLMs focused on analyzing the causal relationship between external inputs and model outputs. This kind of studies~\cite{chuang2025selfciteselfsupervisedalignmentcontext,li2025attributingresponsecontextjensenshannon,cohen2024contextcite} views the model as a black box, without considering its internal inference process, which leads to the skepticism about their reliability. For instance, some studies have pointed out that LLMs can exhibit hallucinated chains of thought~\cite{chen2025reasoningmodelsdontsay,lindsey2025biology}. Therefore, mechanistic interpretability has been proposed, which tries to understand how components of the model function individually, and how they connect and collaborate to enable the model's various behaviors and capabilities. Because they introduce analyses of the model's internal computational process, these studies are considered to reflect the model's behavior more faithfully. The study objects~\cite{rai2024practical} of mechanistic interpretability include the model's components (such as neurons, SAE features, and attention heads), the functional circuits formed among these components, and the universality~\cite{gurnee2024universalneuronsgpt2language} of mechanisms across different models. Discovering mechanisms helps researchers identify which parts of the input or which model components lead to specific behaviors, making the model's overall reasoning process more comprehensible to humans, and thereby addressing potential issues in models such as hallucination, safety, and reliability.

\subsection{SAE Studies}
The application of sparse autoencoders (SAEs) has provided more monosemantic features for models’ internal representations, thereby significantly enhancing interpretability. This has given rise to two main research directions: natural language explanations of SAE features~\cite{bills2023language,paulo2025automaticallyinterpretingmillionsfeatures,lee2023importanceprompttuningautomated} and SAE-based steering. However, the development of natural language explanations faces limitations in two aspects: the inherent ambiguity of natural language itself, and the constraints of fundamental analysis methods. Specifically, for general model behaviors (as opposed to specific concepts~\cite{Wu2025AxBenchSL} or knowledge), accurately capturing their functional patterns solely by analyzing token segments in the activated context is very challenging. Moreover, such analyses are often confined to specific examples and lack generalizability, resulting in studies with limited practical value. In contrast, SAE-based steering methods bypass the step of providing natural language explanations for features. They directly link features to target behaviors through statistical correlations, demonstrating broader applicability in practice.

\subsection{Model Editing}
Studies on model editing can be divided into two categories: editing for specific knowledge and editing for behavioral guidance. Knowledge editing~\cite{li2024pmet,fang2025alphaeditnullspaceconstrainedknowledge} is usually a fine-grained operation, primarily focusing on the feed-forward network modules, which are recognized to be responsible for internal knowledge retrieval. This kind of approach aims to precisely inject or modify specific knowledge while avoiding impacts on other knowledge within the model. In contrast, behavioral editing is a broader method. Its goal is not to alter specific knowledge, but to guide specific behavioral patterns in the model—such as refusing to answer inappropriate questions in security scenarios~\cite{zhou2025roleattentionheadslarge}, or making models' outputs more faithful to external contextual knowledge in RAG tasks with knowledge conflicts.

\section{Prompt Template}
\label{sec:Prompt Template}
\subsection{Question Answering}
When sampling answers, we use the prompt template of question answering shown in Figure~\ref{fig:PromptTemplateQA}. It is a 3-shot prompt template, which aims to control the output format of the models. When we need the closed-book answers to test the parametric knowledge, the contextual parts are excluded. 

When calculating the mutual information, following the method of SPARE~\cite{zhao2025steering}, the few-shot examples with parametric knowledge are sampled by different seeds and numbers. Details of few-shot examples can be seen in Appendix~\ref{sec:implementation-details}.

\subsection{LLM judgement}
We use the prompt template shown in Figure~\ref{fig:PromptTemplateJudge} to help judge the answer by LLM. As we observe some bad cases during judgement, we add some judging examples to improve them.

\section{Implementation Details}
\label{sec:implementation-details}
\subsection{Details of Datasets}
\label{section:C1}
To obtain more accurate examples that represent the knowledge selection behavior, we sample the answers and check their faithfulness when using original context and substituted context. We filter out the bad cases in which the model's answers to substituted context are judged both right when compared with the original and the substituted golden answers. Moreover, the length of the input tokens is limited to 256 to get more consistent steering effect comparison.

\subsection{Calculating Mutual Information}
In the first selection with the fixed prompt shown in Figure~\ref{fig:PromptTemplateQA} by LLM judgement in Section~\ref{section:C1}, we pick out 1,000 / 200 positive and negative examples for Macnoise / NQ-Swap respectively. 
When calculating the mutual information, using different numbers and instances of few-shot examples is beneficial to obtain better results. In order to conduct a fair comparison and fit the provided hyperparameters of the baseline, we use samples from the memorized sets provided by SPARE~\cite{zhao2025steering} for Gemma-2-9B and Llama-3-8B. For Llama-3.1-8B that is not used by the baseline, we follow the construction method using 5 different seeds and numbers of few-shot examples from 3 to 5. For each few-shot setting, we collect the activations of faithful and unfaithful examples for the target and nontarget sets. As the few-shot examples are changed, the positive and negative examples are sampled and evaluated again to ensure the behaviors are not changed, which means the examples are filtered twice. The answers are checked by judging whether they are the substrings of the substituted context for this second filtering. The few-shot examples are sampled in development datasets with 128 questions, and the test examples are from the rest parts.

It should be noted that the few-shot examples need to be without knowledge conflicts, which means the model has the parametric knowledge that is the same as the context. In this way, the model won't learn the knowledge selection tendency from the few-shot examples.

\subsection{Details of KPI}
More sensitivity analysis for steering strength, top-k and number of sampled instances can be seen in Appendix~\ref{Sensitivity}.

We first calculate mutual information on a wide range of layers. We pick 18-36 layers of Gemma-2-9B, 14-23 layers of Llama-3-8B and 14-27 layers of Llama-3.1-8B. Each top-10 (top-k gained by explorations) positive feature of every layer is steered individually to record its top-5 enhanced features of each affected layer. The enhancement check is performed on a small development dataset with the size of 10, because of the relatively stable influence. Based on the enhancement results, we build the interaction graph of the positive features and rank them by the indegree. Among the top indegree-ranked features, we check their single steering effects on a larger development dataset with the size of 200 to obtain the revised counts of the question answering. The best feature with the largest revised count is selected and its layer is regarded as the key layer, which is selected iteratively until there are no better top-ranked features in preceding layers. The key layer selection can also solve the issue of higher indegrees in later layers to help us locate the true functional layers. Then the top indegree-ranked features in the key layer and its preceding layers are selected as the key steering features. We also check their single steering effects and decide the amount of steering features.

The detailed selected results can be seen in the Table~\ref{tab:KPI features Macnoise} and Table~\ref{tab:KPI features NQSwap}. The features are recorded in the form of triples, which respectively represent the layer, the id number and the maximum activation value. The maximum activation value multiplied by the strength coefficient constitutes $\alpha$ in Formula~\ref{steering formula}. For single steering, we use the strength coefficients of 2.0 for Gemma-2-9B, and 3.0 for Llama-3-8B and Llama-3.1-8B, which are gained by explorations on the development datasets. The test datasets for results in Table~\ref{tab:main-results} and Table~\ref{tab:side-results} are with the size of 200. 
\begin{table*}
\centering
\small
\setlength\tabcolsep{10pt}
\begin{tabular}{m{1.8cm}m{1.2cm}m{8.5cm}m{1.5cm}}
\toprule
 Model & Key Layer & All Steering Features & Strength Coefficient\\  
\midrule
Gemma-2-9B & 28 &(28,84485,60.297),(24,76071,33.874),(25,39999,21.441),
(25,77008,38.481),(27,644,47.593),(26,7425,26.511),
(27,123319,31.519),(28,9032,88.5) & 0.9
\\ \midrule
Llama-3-8B & 18 & (17,89349,1.471),(18,52958,2.166),(18,61143,1.606),
(17,126823,1.484),(17,71518,0.921),(18,30899,1.77),
(16,126464,1.154),(18,111786,0.826),(17,113473,1.767),
(17,39250,1.289)
& 1.2
\\ \midrule
Llama-3.1-8B & 23 &(23,9812,5.5),(21,25813,6.188),(21,7884,5),
(23,103034,4.938),(20,1850,3.281),(18,3799,3.734),
(22,31181,5.688),(21,115719,5.875),(22,114544,3.234),
(22,44606,3.453)& 1.5 \\
\bottomrule
\end{tabular}
\caption{KPI selection results of each model on Macnoise.}
\label{tab:KPI features Macnoise}
\end{table*}

\begin{table*}
\centering
\small
\setlength\tabcolsep{10pt}
\begin{tabular}{m{1.8cm}m{1.2cm}m{8.5cm}m{1.5cm}}
\toprule
 Model & Key Layer & All Steering Features & Strength Coefficient\\  
\midrule
Gemma-2-9B & 28 & (28,84485,60.297),(26,52871,42.875),(26,7425,26.511),
(25,77008,38.481),(24,76071,33.874),(23,116937,56.364),
(25,39999,21.441),(23,32901,23.672),(26,810,31.104),
(25,3118,31.488),(28,9032,88.5)& 0.9
\\ \midrule
Llama-3-8B & 18 &(18,52958,2.266),(17,126823,1.415),(17,89349,1.358),
(16,22835,0.979),(18,111786,0.979),(17,71518,0.826),
(15,27159,1.134),(17,43840,1.266),(15,93728,0.824),
(17,88596,1.16),(16,65575,0.807),(17,111193,0.956),
(18,80716,0.699),(17,39250,1.348),(16,121234,1.258)
& 0.9
\\ \midrule
Llama-3.1-8B & 23 & (23,9812,5.5),(21,25813,6.188),(23,24626,5.938),
(21,118917,2.516),(21,81013,3),(20,1850,3.281),
(21,48632,5.656),(19,6817,2.719),(21,130035,4.656),
(23,50751,7.625) & 1.3 \\
\bottomrule
\end{tabular}
\caption{KPI selection results of each model on NQ-Swap.}
\label{tab:KPI features NQSwap}
\end{table*}

\subsection{Details of Baselines}

\paragraph{ICL}\citep{GPT3}: We find different few-shot examples have quite different abilities to guide the model. Therefore, using 5 seeds, we select 3 examples that show faithful behavior with knowledge conflicts as the few-shot examples and report the best performance.

\paragraph{CAD}\citep{shi-etal-2024-trusting}: The only hyperparameter is the combination coefficient $\alpha$. We test it ranging from 0.1 to 1.5 with an interval of 0.1. We finally set it to 0.5 for all the models based on explorations.
\paragraph{SPARE}\citep{zhao2025steering}: We follow the hyperparameters $K$ specified in the original paper for Gemma-2-9B and Llama-3-8B, and make explorations for $\alpha$. All the hyperparameters for Llama-3.1-8B are explored by following the baseline setting. Specifically, the hyperparameters $K$ and $\alpha$ are set to 0.01 and 3 for Gemma-2-9B and Llama-3.1-8B, and 0.07 and 2 for Llama-3-8B. The steering layers are as follows: layers 23, 24, 25, 29, 30 and 31 for Gemma-2-9B, and layers 13-16 for both Llama-3-8B and Llama-3.1-8B. SPARE uses a remove-and-add steering operation. We do not include the removal operation in our method for two reasons: (1) in our preliminary exploration, it was less effective and harder to control; and (2) it is not central to the design of our graph-based method. Moreover, even without removal, our refined add-only method outperforms the remove-and-add baseline, suggesting that the main benefits of our approach come from the graph-based selection design.

\paragraph{STA}\citep{bricken2023monosemanticity}: Since the original paper does not propose a specific method for layer selection apart from exploration, we select the same steering layers as those used in SPARE to evaluate the performance differences of different feature identification methods. The number of positive and negative examples used for selection is the same as that used for calculating mutual information, specifically 1,000 for Macnoise and 200 for NQ-Swap. Based on exploration, the steering strength coefficient $\lambda$ for Macnoise is 20.0 for Gemma-2-9B, and 2.0 for both Llama-3-8B and Llama-3.1-8B. For NQ-Swap, $\lambda$ is set to 65.0 for Gemma-2-9B, and 2.0 for both Llama-3-8B and Llama-3.1-8B. The amplitude threshold $\alpha$ and frequency threshold $\beta$ are both 0.35.

\begin{figure}[t]
\centering
\includegraphics{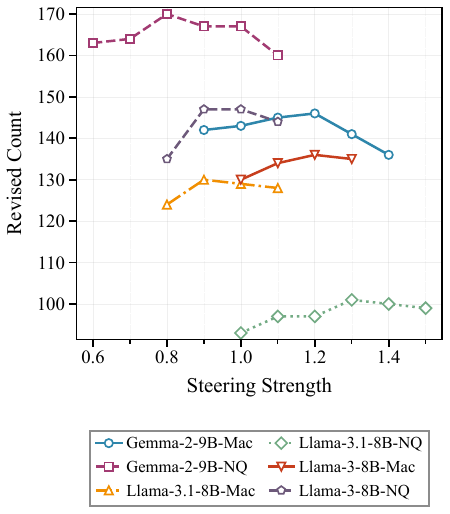}
  \caption{Explorations of steering strength for different models and datasets.}
  \label{fig:steering strength}
\end{figure}

\section{Results in More Complex Faithful Scenario}
\label{complex-faithful}
To test the generalizability of our method in improving faithfulness, we choose FaithEval-counterfactual-v1.0~\cite{ming2024faitheval} as supplement. It has completely designed contexts for the counterfactual answers, which are more complex than entity replacement in the two original datasets. As it has no factual context and the corresponding answer, we test whether using the steering settings derived from the two simple datasets can improve the performance in this scenario, which gives more challenges for unseen real-world knowledge conflicts.

Here we present the results in Table~\ref{tab:counter-results}. Notice that, at this time, we use all the dataset for testing without further filtering or division. The results of our method show the consistent and significant improvement, verifying the generalizability of our method for out-of-distribution knowledge conflicts.

\begin{table*}[ht!]
\centering
\small
\resizebox{\linewidth}{!}{
\begin{tabular}{lcccccc} 
\toprule
\bf Method & \multicolumn{3}{c}{\textbf{NQ-Swap}} & \multicolumn{3}{c}{\textbf{Macnoise}} \\ 
\cmidrule(lr){2-4} \cmidrule(lr){5-7}
& \bf Llama-3-8B & \bf Llama-3.1-8B & \bf Gemma-2-9B & \bf Llama-3-8B & \bf Llama-3.1-8B & \bf Gemma-2-9B \\
\midrule
Without Steering & $-$ & $-$ & $-$ & $51.7$ & $48.0$ & $52.0$ \\
SPARE & $22.4$ & $13.1$ & $50.6$ & $20.0$ & $11.1$ & $60.4$ \\
KPI & $\textbf{58.1}$ & $\textbf{60.9}$ & $\textbf{60.5}$ & $\textbf{61.2}$ & $\textbf{60.2}$ & $\textbf{61.4}$ \\

\bottomrule
\end{tabular}
}
\caption{Results in a more complex faithful scenario of FaithEval-counterfactual-v1.0. Our method has the consistent and significant improvement, showing the generalizability of our method for out-of-distribution knowledge conflicts. Results without steering are the same in NQ-Swap and Macnoise.}
\label{tab:counter-results}
\end{table*}

\section{Side Effect Results on Other Model Capabilities}
\label{Other Model Capabilities}
Apart from the side effect experiments in the original context-understanding task, here we present the results for other model capabilities. We choose MMLU~\cite{hendrycks2020measuring} and GSM8K~\cite{cobbe2021training} for investigation.

We use the test code of EleutherAI\footnote{https://github.com/EleutherAI/lm-evaluation-harness}, and fix its steering positions of all tokens in the original setting to the last token in the prompt, which is consistent with our setting. We use the one-shot setting for both the datasets. And for GSM8K, results are reported on flexible-extracted answers with exact match. The results of steering on the last token in the prompt can be seen in Table~\ref{tab:mmlu-side-results-last} and Table~\ref{tab:gsm8k-side-results}. Our method shows comparable results on MMLU and consistently better results than the baseline on GSM8K. For a more obvious comparison, we report the results of steering on all tokens on MMLU (Table~\ref{tab:mmlu-side-results-all}), the performances of the best baseline settings  collapse in the same way that they output some formats of URLs (for example, www.://). And our method shows small side effects, which maintains a high level of capability.

\begin{table*}[ht!]
\centering
\small
\resizebox{\linewidth}{!}{
\begin{tabular}{lcccccc} 
\toprule
\bf Method & \multicolumn{3}{c}{\textbf{NQ-Swap}} & \multicolumn{3}{c}{\textbf{Macnoise}} \\ 
\cmidrule(lr){2-4} \cmidrule(lr){5-7}
& \bf Llama-3-8B & \bf Llama-3.1-8B & \bf Gemma-2-9B & \bf Llama-3-8B & \bf Llama-3.1-8B & \bf Gemma-2-9B \\
\midrule
Without Steering & $-$ & $-$ & $-$ & $64.08_{\pm 0.38}$ & $64.36_{\pm 0.38}$ & $69.57_{\pm 0.36}$ \\
SPARE & $64.11_{\pm 0.38}$ & $64.31_{\pm 0.38}$ & $68.89_{\pm 0.36}$ &  $64.24_{\pm 0.38}$ & $64.27_{\pm 0.38}$ & $68.88_{\pm 0.36}$ \\
KPI & $64.11_{\pm 0.38}$ & $64.34_{\pm 0.38}$
 & $69.56_{\pm 0.36}$ & $64.19_{\pm 0.38}$ & $64.18_{\pm 0.38}$ & $69.58_{\pm 0.36}$  \\

\bottomrule
\end{tabular}
}

\caption{Side effect checks (steering on the last token in prompt) of our method on MMLU datasets. Our method achieves comparable performance, showing its small side effects in other inference scenarios of universal knowledge.}
\label{tab:mmlu-side-results-last}
\end{table*}

\begin{table*}[ht!]
\centering
\small
\resizebox{\linewidth}{!}{
\begin{tabular}{lcccccc} 
\toprule
\bf Method & \multicolumn{3}{c}{\textbf{NQ-Swap}} & \multicolumn{3}{c}{\textbf{Macnoise}} \\ 
\cmidrule(lr){2-4} \cmidrule(lr){5-7}
& \bf Llama-3-8B & \bf Llama-3.1-8B & \bf Gemma-2-9B & \bf Llama-3-8B & \bf Llama-3.1-8B & \bf Gemma-2-9B \\
\midrule
Without Steering & $-$ & $-$ & $-$ & $38.21_{\pm 1.34}$ & $40.94_{\pm 1.35}$ & $62.62_{\pm 1.33}$ \\
SPARE & $31.99_{\pm 1.28}$ & $36.69_{\pm 1.33}$ & $55.42_{\pm 1.37}$ &  $33.06_{\pm 1.30}$ & $37.76_{\pm 1.34}$ & $52.99_{\pm 1.37}$ \\
KPI & $32.98_{\pm 1.29}$ & $38.82_{\pm 1.34}$ & $61.56_{\pm 1.34}$ &  $35.71_{\pm 1.32}$ & $37.83_{\pm 1.34}$ & $61.03_{\pm 1.34}$ \\

\bottomrule
\end{tabular}
}
\caption{Side effect checks (steering on the last token in prompt) of our method on GSM8K datasets. Our method achieves better performance than the mass steering baseline, showing its ability of alleviating side effects in other inference scenarios of math.}
\label{tab:gsm8k-side-results}
\end{table*}

\begin{table*}[ht!]
\centering
\small
\resizebox{\linewidth}{!}{
\begin{tabular}{lcccccc} 
\toprule
\bf Method & \multicolumn{3}{c}{\textbf{NQ-Swap}} & \multicolumn{3}{c}{\textbf{Macnoise}} \\ 
\cmidrule(lr){2-4} \cmidrule(lr){5-7}
& \bf Llama-3-8B & \bf Llama-3.1-8B & \bf Gemma-2-9B & \bf Llama-3-8B & \bf Llama-3.1-8B & \bf Gemma-2-9B \\
\midrule
Without Steering & $-$ & $-$ & $-$ & $64.08_{\pm 0.38}$ & $64.36_{\pm 0.38}$ & $69.57_{\pm 0.36}$ \\
SPARE & $24.21_{\pm 0.36}$ & $26.31_{\pm 0.37}$ & $22.95_{\pm 0.35}$ &  $24.17_{\pm 0.36}$ & $30.09_{\pm 0.39}$ & $22.95_{\pm 0.35}$ \\
KPI & $63.39_{\pm 0.38}$ & $63.47_{\pm 0.38}$
 & $60.13_{\pm 0.39}$ & $63.25_{\pm 0.38}$ & $62.53_{\pm 0.39}$ & $67.12_{\pm 0.37}$  \\

\bottomrule
\end{tabular}
}

\caption{Side effect checks (steering on all tokens) of our method on MMLU datasets. Our method maintains a high level of capability, showing its ability of alleviating side effects by filtering redundant features.}
\label{tab:mmlu-side-results-all}
\end{table*}

\section{Sensitivity Analysis}
\label{Sensitivity}
\subsection{Exploration of Steering Strength}
Here we present part of our exploration results of steering strength coefficients in Figure~\ref{fig:steering strength}. In some settings of the main results, we do not use the best coefficients, because they show relatively lower performance in the side effect checks. We can observe that the revised count shows a roughly single-peaked pattern depending on the steering strength. When the steering strength is too small, the steering vector does not work for the best. And when the steering strength is too large, it might compromise the model's ability to respond normally. The dynamic regulation of steering strength is studied in some other works~\cite{li2026efficient}, which is not the focus of our work.

\subsection{Exploration of Top-k}
\begin{table}[htbp]
\centering

\begin{tabular}{lcccc}
\toprule
Top-k & 10 & 15 & 20 & 30 \\
\hline
Gemma-2-9B & 1.0 & 0.8 & 0.8 & 0.7 \\
Llama-3-8B & 1.0 & 0.7 & 0.7 & 0.6 \\
\bottomrule
\end{tabular}
\caption{The degree of overlap of selected features with different top-k settings.}
\label{tab:overlap-topk}
\end{table}

\begin{table}[htbp]
\centering

\begin{tabular}{lccccc}
\toprule
Top-k & 5 & 10 & 15 & 20 & 30 \\
\hline
Gemma-2-9B & 148 & 150 & 140 & 146 & 135 \\
Llama-3-8B & 114 & 141 & 139 & 136 & 132 \\
\bottomrule
\end{tabular}
\caption{The performance (revised count) changes of selected features with different top-k settings.}
\label{tab:performance-changes-topk}
\end{table}

Here we present the degree of overlap of selected features (top-10) for different layer top-k settings in Table~\ref{tab:overlap-topk}. It shows stability that a majority of important features keep selected.

Here we present the performance changes of selected features with different layer top-k settings in Table~\ref{tab:performance-changes-topk} (feature amounts are set to 10 for Gemma-2-9B and 15 for Llama-3-8B respectively, and other hyperparameters remain the same). Gemma-2-9B is tested on Macnoise, and Llama-3-8B is tested on NQ-Swap. When the top-k is set as 10, it shows the best performances, excluding a large amount of noise features and including the potentially good ones. The results also show that overly aggressive pruning by top-k can remove useful features and impair the performance, which is consistent with our analysis in Section~\ref{sec:feature-redundancy}.

\subsection{Stability of Feature Interaction Pattern Capture with Different Numbers of Sampled Instances}
Here we present the degree of overlap of selected features (top-15) with different numbers of sampled instances in Table~\ref{tab:overlap-num-instance}. Gemma-2-9B is tested on Macnoise, and Llama-3-8B is tested on NQ-Swap. 

The results show that the interaction patterns captured by our single steering method are quite stable with the number of sampled instances. So, the small development datasets used in our main experiments are reasonable.

\begin{table}[htbp]
\centering
\begin{tabular}{lccc}
\toprule
Number of Instances & 10 & 100 & 200 \\
\hline
Gemma-2-9B & 1.0 & 0.8  & 0.73  \\
Llama-3-8B & 1.0 & 0.87  & 0.87  \\
\bottomrule
\end{tabular}
\caption{The degree of overlap of selected features with different numbers of sampled instances.}
\label{tab:overlap-num-instance}
\end{table}

\section{Efficiency Analysis}
To estimate the interactions among features, the method of backward propagation calculates the gradients layer by layer, which requires substantial GPU computing power and storage space due to the enormous number of SAE features. Our method uses forward passes with single steering to significantly improve the efficiency. The time savings come from leveraging prior knowledge of mutual information, which allows us to identify the important features. The reason why methods like Circuit Tracer~\cite{ameisen2025circuit} use backward propagation is that they rely on this process to locate the important features, but it also has the drawbacks that are discussed in~\nameref{limitation}.

To better quantify the computational cost of graph construction, here we report results for Gemma-2-9B, which is the largest model in our experiments. All experiments are conducted on a single NVIDIA A100-SXM4-40GB GPU. The maximum GPU memory usage is 36.45 GiB. Evaluation follows the default experimental setting of top-10 features per layer, recording top-5 enhanced features per downstream layer, a development set of size 10 and analyzing layers 18 to 36. We report layer-wise time cost in Table~\ref{tab:layer_time_cost}, because later source layers have fewer downstream layers to analyze. The total computation cost is 3.89 GPU hours.

\begin{table}[t!]
\centering
\small
\setlength{\tabcolsep}{6pt}
\begin{tabular}{cc}
\toprule
Layer & Runtime (s) \\
\midrule
18 & 1535.77 \\
19 & 1750.06 \\
20 & 1378.83 \\
21 & 1053.75 \\
22 & 1233.42 \\
23 & 882.42 \\
24 & 1152.39 \\
25 & 640.63 \\
26 & 721.01 \\
27 & 636.88 \\
28 & 554.89 \\
29 & 508.84 \\
30 & 472.61 \\
31 & 392.92 \\
32 & 338.13 \\
33 & 248.00 \\
34 & 202.31 \\
35 & 180.56 \\
36 & 105.12 \\
\bottomrule
\end{tabular}
\caption{Per-layer runtime in seconds for layers 18 to 36 on Gemma-2-9B under the default experimental setting.}
\label{tab:layer_time_cost}
\end{table}

\begin{figure*}[t]
    \centering
    \begin{tcolorbox}[
        colback=green!5,
        colframe=green!35!black,
        boxrule=0.5pt,
        title={\textbf{Prompt Template of Question Answering}},
        width=\linewidth
    ]
Context:Albert Einstein developed the theory of relativity in 1905\\Question:Who developed the theory of relativity?\\Answer:Albert Einstein\\\\Context:Paris is the capital of France\\Question:What is the capital of France?\\Answer:Paris\\\\Context:Alexander Graham Bell invented the telephone in 1876\\Question:Who invented the telephone?\\Answer:Alexander Graham Bell\\\\Context:…\\Question:…\\Answer:
    \end{tcolorbox}
    \caption{Prompt template of question answering. The last colon is where we perform activation sampling.}
    \label{fig:PromptTemplateQA}
    \vspace{-15pt}
\end{figure*}

\begin{figure*}[t]
    \centering
    \begin{tcolorbox}[
        colback=green!5,
        colframe=green!35!black,
        boxrule=0.5pt,
        title={\textbf{Prompt Template of LLM Judgement}},
        width=\linewidth
    ]
You are an answer evaluator. Based on the question and the golden answer, you should judge a given answer is right or wrong.\\
If it is right, only output '1', or only output '0'.\\
Attention! Only judge the candidate answer as wrong when it is highly different from the golden answer.\\
For example:\\
1:When the golden answer is "March 28, 1941", and the candidate answer is "1941", it should be judged as right.\\
2:When the golden answer is "16-year-old", and the candidate answer is "16", it should be judged as right.\\
3:When the golden answer is "Stylianos "Stelios" Kyriakides", and the candidate answer is "Stylianos Kyriakides", it should be judged as right.\\
The question is: …\\
The golden answer is: …\\
The answer that you should evaluate is: …\\
Output: 
    \end{tcolorbox}
    \caption{Prompt template of LLM judgement.}
    \label{fig:PromptTemplateJudge}
    \vspace{-15pt}
\end{figure*}

\section{Formalization of Key Path Identification}
\label{app:kpi}

Let $\mathcal{L}$ denote the set of selected transformer layers. For each
$\ell \in \mathcal{L}$, let $\mathcal{F}_{\ell}$ be the set of all SAE features
at layer $\ell$, and let $\mathcal{V}_{\ell}\subseteq \mathcal{F}_{\ell}$ be the
set of candidate positive features selected at layer $\ell$. We write $\ell(v)$
for the layer index of feature $v$, and define the overall candidate feature set as
\[
\mathcal{V}= \bigcup_{\ell\in\mathcal{L}} \mathcal{V}_{\ell}.
\]

\paragraph{Interaction graph}
We define the feature interaction graph as an unweighted directed graph
$G=(\mathcal{V},\mathcal{E})$, where each node corresponds to one candidate feature.
Let $\mathcal{D}_{\mathrm{graph}}$ be a small dataset used for graph construction.
Since steering a feature can only affect representations at the same layer or
later layers, for a source feature $u\in\mathcal{V}$, candidate target features
are restricted to layers $\ell' \ge \ell(u)$. For any such layer $\ell'$ and any
feature $v\in\mathcal{F}_{\ell'}$, let $a_v(x)$ denote the activation of $v$ on
example $x$, and let $a_v^{(u)}(x)$ denote the activation of $v$ when only feature
$u$ is steered. We define the mean activation increase from $u$ to $v$ as
\[
\bar{\Delta}_{u\to v}
=
\frac{1}{|\mathcal{D}_{\mathrm{graph}}|}
\sum_{x\in\mathcal{D}_{\mathrm{graph}}}
\bigl(a_v^{(u)}(x)-a_v(x)\bigr).
\]
For each source feature $u$ and each candidate layer $\ell' \ge \ell(u)$, we define
\[
\mathrm{Top5}_{\ell'}(u)
=
\operatorname{Top5}\bigl(\mathcal{F}_{\ell'};\bar{\Delta}_{u\to \cdot}\bigr)
\cap \mathcal{V}_{\ell'},
\]
where $\operatorname{Top5}(\mathcal{F}_{\ell'};\bar{\Delta}_{u\to \cdot})$ denotes
the five features in $\mathcal{F}_{\ell'}$ with the largest positive mean activation
increases under steering feature $u$. The edge set is then defined as
\[
\mathcal{E}
=
\{(u,v)\mid v\in \bigcup_{\ell'\ge \ell(u)} \mathrm{Top5}_{\ell'}(u)\}.
\]

\paragraph{Key feature ranking}
We score each feature by its indegree in the interaction graph:
\[
\deg^{\mathrm{in}}(v)
=
\sum_{u\in\mathcal{V}} \mathbf{1}[(u,v)\in\mathcal{E}].
\]
We then rank all features in $\mathcal{V}$ jointly by decreasing indegree, and
denote the resulting global ranking by $\mathcal{R}$.

\paragraph{Key layer selection and key path construction}
Let $\mathcal{D}_{\mathrm{eval}}$ be a larger validation set used for layer selection.
Let
\[
\mathcal{R}^{(r)}
=
\operatorname{TopR}(\mathcal{V};\mathcal{R})
\]
denote the top-$r$ features in the global ranking $\mathcal{R}$. For each layer $\ell$,
we then define
\[
\mathcal{S}_{\ell}
=
\mathcal{R}^{(r)} \cap \mathcal{V}_{\ell}.
\]
For any candidate key layer $\ell^\star\in\mathcal{L}$, we define the corresponding
key-path feature set as
\[
\mathcal{P}_{\ell^\star}
=
\bigcup_{\ell\le \ell^\star}\mathcal{S}_{\ell}.
\]
We select the key layer by
\[
\hat{\ell}
=
\arg\max_{\ell^\star\in\mathcal{L}}
\mathrm{Eval}(M,\mathcal{S}_{\ell^\star},\mathcal{D}_{\mathrm{eval}}),
\]
where $\mathrm{Eval}(M,\mathcal{S},\mathcal{D}_{\mathrm{eval}})$ denotes the best steering
performance obtained by singly steering each feature in $\mathcal{S}$ on
$\mathcal{D}_{\mathrm{eval}}$. The final key path is defined as
\[
\mathcal{P}=\mathcal{P}_{\hat{\ell}}.
\]

\paragraph{Steering strength coefficient tuning}
We tune the steering strength coefficient over a candidate set $\mathcal{A}$ by
\[
\alpha^\star
=
\arg\max_{\alpha\in\mathcal{A}}
\mathrm{Eval}(M,\mathcal{P},\alpha,\mathcal{D}_{\mathrm{eval}}),
\]
where $\mathrm{Eval}(M,\mathcal{P},\alpha,\mathcal{D}_{\mathrm{eval}})$ denotes the validation
performance when steering all features in $\mathcal{P}$ with strength coefficient $\alpha$.

\begin{algorithm*}[t]
\caption{Key Path Identification (KPI)}
\label{alg:kpi}
\begin{algorithmic}[1]
\Require model $M$, selected layers $\mathcal{L}$, all-feature sets $\{\mathcal{F}_{\ell}\}_{\ell\in\mathcal{L}}$, candidate feature sets $\{\mathcal{V}_{\ell}\}_{\ell\in\mathcal{L}}$, graph-construction set $\mathcal{D}_{\mathrm{graph}}$, evaluation set $\mathcal{D}_{\mathrm{eval}}$, global ranking budget $r$, strength coefficient set $\mathcal{A}$
\State $\mathcal{V}\gets\bigcup_{\ell\in\mathcal{L}}\mathcal{V}_{\ell}$, $\mathcal{E}\gets\emptyset$
\ForAll{$u\in\mathcal{V}$}
    \ForAll{$\ell' \in \mathcal{L}$ such that $\ell' \ge \ell(u)$}
        \State compute $\bar{\Delta}_{u\to v}$ for all $v\in\mathcal{F}_{\ell'}$ using $\mathcal{D}_{\mathrm{graph}}$
        \State $\mathrm{Top5}_{\ell'}(u)\gets \operatorname{Top5}(\mathcal{F}_{\ell'};\bar{\Delta}_{u\to\cdot}) \cap \mathcal{V}_{\ell'}$
        \State add $(u,v)$ to $\mathcal{E}$ for all $v\in \mathrm{Top5}_{\ell'}(u)$
    \EndFor
\EndFor
\State compute $\deg^{\mathrm{in}}(v)$ for all $v\in\mathcal{V}$
\State rank all features in $\mathcal{V}$ jointly by decreasing indegree to obtain $\mathcal{R}$
\State $\mathcal{V}^{(r)} \gets \operatorname{TopR}(\mathcal{V};\mathcal{R})$
\ForAll{$\ell\in\mathcal{L}$}
    \State $\mathcal{S}_{\ell}\gets \mathcal{V}^{(r)} \cap \mathcal{V}_{\ell}$
\EndFor
\State $\hat{\ell}\gets \arg\max_{\ell^\star\in\mathcal{L}} \mathrm{Eval}(M,\mathcal{S}_{\ell^\star},\mathcal{D}_{\mathrm{eval}})$
\State $\mathcal{P}\gets \bigcup_{\ell\le \hat{\ell}}\mathcal{S}_{\ell}$
\State $\alpha^\star\gets \arg\max_{\alpha\in\mathcal{A}} \mathrm{Eval}(M,\mathcal{P},\alpha,\mathcal{D}_{\mathrm{eval}})$
\State \Return $\mathcal{P}, \alpha^\star$
\end{algorithmic}
\end{algorithm*}

\end{document}